\documentclass[pdflatex,sn-mathphys-num]{sn-jnl}

\usepackage{graphicx}%
\usepackage{multirow}%
\usepackage{amsmath,amssymb,amsfonts}%
\usepackage{amsthm}%
\usepackage{mathrsfs}%
\usepackage[title]{appendix}%
\usepackage{xcolor}%
\usepackage{textcomp}%
\usepackage{manyfoot}%
\usepackage{booktabs}%
\usepackage{algorithm}%
\usepackage{algorithmicx}%
\usepackage{algpseudocode}%
\usepackage{listings}%

\usepackage[caption=false]{subfig}%
\usepackage[singlelinecheck=false, font=small,labelfont=bf,
   justification=justified,
   format=plain]{caption}
\usepackage{adjustbox}
\theoremstyle{thmstyleone}%
\theoremstyle{thmstyletwo}%

\theoremstyle{thmstylethree}%

\newcommand{\parref}[1]{(\ref{#1})}

\begin{document}

\title[Article Title]{\textbf{KNN-Defense: Defense against 3D Adversarial Point Clouds using Nearest-Neighbor Search}}

\author[1]{\fnm{Nima} \sur{Jamali}}\email{nima.jamali@uwaterloo.ca}

\author[1]{\fnm{Matina} \sur{Mahdizadeh Sani}} \email{m3mahdiz@uwaterloo.ca}

\author*[2]{\fnm{Hanieh} \sur{Naderi}}\email{hanieh.naderi@ut.ac.ir}

\author*[3]{\fnm{Shohreh} \sur{Kasaei}}\email{kasaei@sharif.edu}

\affil[1]{\orgdiv{\normalsize School of Computer Science}, \orgname{University of Waterloo}}

\affil[2]{\orgdiv{\normalsize Department of Data Science and Technology, School of Intelligent Systems Engineering}, \orgname{University of Tehran}}

\affil[3]{\orgdiv{\normalsize Department of Computer Engineering}, \orgname{Sharif University of Technology}}


\abstract{

Deep neural networks (DNNs) have demonstrated remarkable performance in analyzing 3D point cloud data. However, their vulnerability to adversarial attacks—such as point dropping, shifting, and adding—poses a critical challenge to the reliability of 3D vision systems. These attacks can compromise the semantic and structural integrity of point clouds, rendering many existing defense mechanisms ineffective.
To address this issue, a defense strategy named KNN-Defense is proposed, grounded in the manifold assumption and nearest-neighbor search in feature space. Instead of reconstructing surface geometry or enforcing uniform point distributions, the method restores perturbed inputs by leveraging the semantic similarity of neighboring samples from the training set. KNN-Defense is lightweight and computationally efficient, enabling fast inference and making it suitable for real-time and practical applications.
Empirical results on the ModelNet40 dataset demonstrated that KNN-Defense significantly improves robustness across various attack types. In particular, under point-dropping attacks—where many existing methods underperform due to the targeted removal of critical points—the proposed method achieves accuracy gains of 20.1\%, 3.6\%, 3.44\%, and 7.74\% on PointNet, PointNet++, DGCNN, and PCT, respectively. These findings suggest that KNN-Defense offers a scalable and effective solution for enhancing the adversarial resilience of 3D point cloud classifiers. (An open-source implementation of the method, including code and data, is available at \href{https://github.com/nimajam41/3d-knn-defense}{https://github.com/nimajam41/3d-knn-defense}).}




\maketitle

\section{Introduction}\label{sec1}

Deep neural networks (DNNs) have achieved remarkable success in a wide range of machine learning tasks, notably in image classification \cite{he2016deep, tan2019efficientnet, dosovitskiy2020image} and image segmentation \cite{chen2018encoder, xie2021segformer, chen2022vision}.
Despite these successes, they remain highly vulnerable to adversarial examples \cite{szegedy2014intriguing}.
These examples consist of inputs subtly modified in a way that deceive the model into incorrect predictions, without perceptible changes to the input itself.

Several 2D adversarial example generation algorithms \cite{goodfellow2015explaining,naderi2021generating,papernot2016limitations,madry2018towards,moosavi2016deepfool,carlini2017towards,he2018decision,su2019one,rahmati2020geoda} have been presented over the years. These algorithms can generally be categorized according to the level of access of the attacker and the techniques used. From these perspectives, adversarial attacks can be classified into different groups, including white-box and black-box attacks, gradient-based and optimization-based attacks, as well as targeted and untargeted attacks.

In a white-box attack scenario, the adversary has complete access to the architecture, parameters, and gradients of the target model. Gradient-based attacks, like FGSM (Fast Gradient Sign Method) \cite{goodfellow2015explaining}, JSMA (Jacobian Saliency Map Attack) \cite{papernot2016limitations}, and PGD (Projected Gradient Descent) \cite{madry2018towards}, use the model’s gradients to generate adversarial examples. Optimization-based attacks, such as DeepFool \cite{moosavi2016deepfool} and C \& W \cite{carlini2017towards}, try to minimize a specific objective function to create adversarial examples. Decision-based attacks, such as DBA \cite{he2018decision}, focus on optimizing directly along the decision boundary of the victim model. All of these methods are considered white-box attacks. In contrast, in a black-box attack scenario, the attacker has limited or no access to the target model's architecture or parameters, and alternative approaches must be taken to generate adversarial examples \cite{su2019one,rahmati2020geoda}. Both white-box and black-box attacks can be further classified into targeted and untargeted attacks. In targeted attacks, the attacker aims to deceive the model into classifying the adversarial example as a certain target class. In contrast, untargeted attacks attempt to cause misclassification without targeting a specific class.

LiDAR sensors have been a highly trending topic of interest recently for their ability to generate highly accurate and dense 3D representations of the surrounding objects. They are widely used in various fields, such as autonomous driving \cite{zhou2018voxelnet, shi2019pointrcnn} and robotics \cite{mei2023overlap, qin2022geometric}. As their popularity continues to grow, researchers have increasingly focused on studying point clouds, the direct output of these sensors. Point clouds are 3D point sets without a specific order used to represent the shapes of real objects. While a very important res ce in many applications, their irregular data format make poses a principal challenge when it comes to processing them using traditional DNNs.  

To address this problem, PointNet \cite{qi2017pointnet} presented an architecture that could handle the issue of unordered sets, while remaining invariant to different transformations. Ever since PointNet has been introduced, 3D point cloud classification has seen numerous advancements. Researchers have developed various architectures, such as PointNet++ \cite{qi2017pointnet++}, DGCNN \cite{wang2019dynamic}, PCT \cite{guo2021pct}, and CurveNet \cite{xiang2021walk} which achieve higher accuracy in classification tasks. 

Recent studies have revealed that point cloud DNNs are also vulnerable to adversarial attacks \cite{xiang2019generating, zheng2019pointcloud, hamdi2020advpc,naderi2022model,arya2021adversarial,naderi2023adversarial}. Similar to 2D attacks, 3D point cloud attacks can be broadly categorized using the same classification schemes, including white-box and black-box, gradient-based and optimization-based, as well as targeted and untargeted attacks. However, what makes 3D attacks unique is the emergence of new types of attacks based on the intrinsic properties of point clouds. In the context of 3D point cloud attacks, due to their set structure, attacks on 3D point clouds can be categorized into three groups: Point-Shifting Attacks, which shift the coordinates of existing points to produce new samples; Point-Generation Attacks, which add new points to the set to induce misclassification; and Point-Dropping Attacks, which remove some points from the set to create adversarial examples. Some of these attacks are capable of producing constructible adversarial examples \cite{tsai2020robust, cao2021invisible}, which pose a threat to many safety-critical applications of 3D point clouds.

To make these networks robust against 3D adversarial point clouds, several methods have been proposed in recent years \cite{liu2019extending, zhou2019dup,wu2020if,naderi2023lpf}. One such method \cite{liu2019extending} uses a combination of original data and adversarial examples to train the model. Another approach is taken by DUP-Net \cite{zhou2019dup}, which leverages statistical outlier removal (SOR) to remove outlier points added or shifted by adversarial attacks, and then feeds the cleaned point cloud data to an upsampling network \cite{yu2018pu}. IF-defense \cite{wu2020if} is another method that uses deep implicit functions to reconstruct the original samples from adversarial ones. These methods primarily focus on surface reconstruction \cite{wu2020if,zhou2019dup,liu2019extending} and maintaining a uniform distribution \cite{wu2020if,zhou2019dup}, which may not be ideal for dropping attacks. The proposed KNN-Defense addresses this limitation by prioritizing the preservation of the point cloud manifold structure, resulting in a more effective defense against various types of attacks, especially dropping attacks.

\begin{figure}[t!]
\centering
\includegraphics[width=0.75\textwidth]{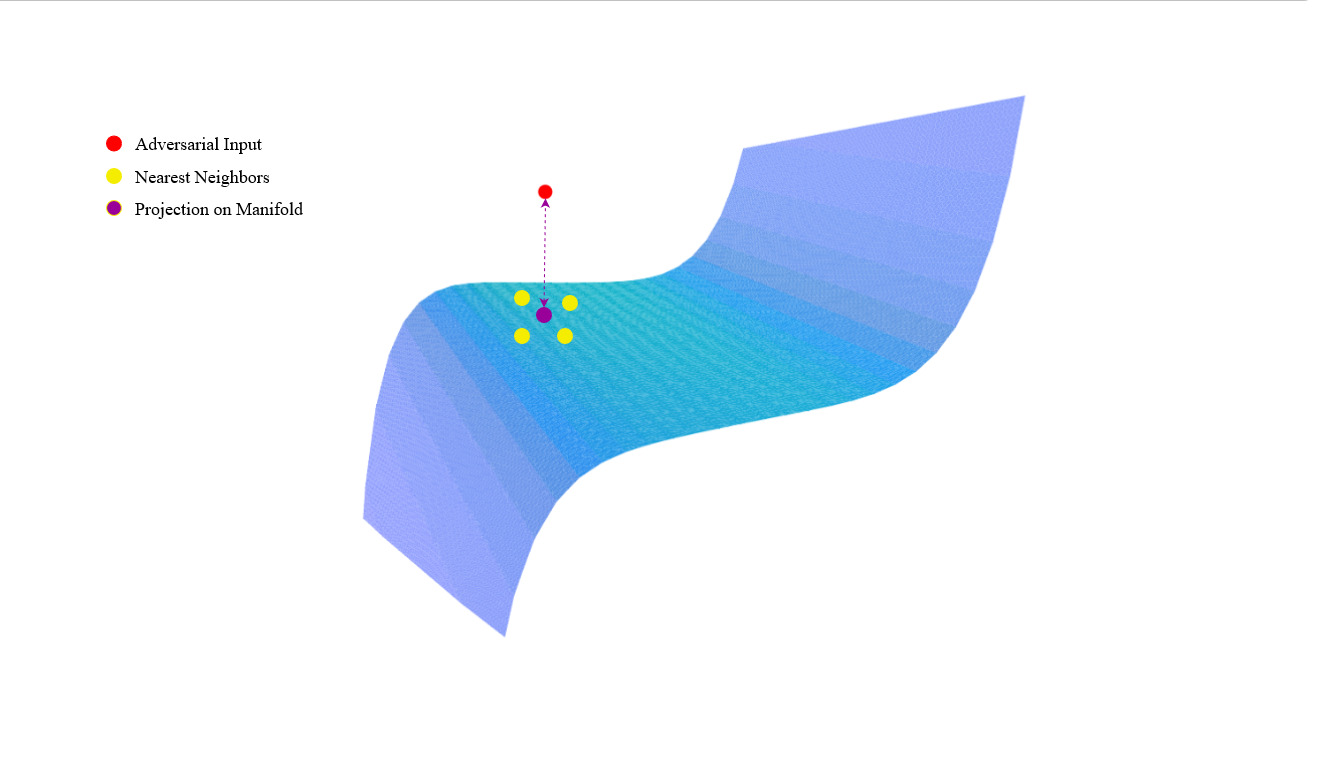}
\caption{Manifold assumption in KNN-Defense. Clean point clouds are assumed to lie on an $n$-dimensional manifold, while adversarial examples deviate from it. The proposed method projects adversarial inputs back onto the manifold for robust classification, using nearest neighbors.}

\label{manifold}
\end{figure}

This work introduces a defense algorithm, KNN-Defense, that leverages the manifold assumption and semantic neighborhood consistency in feature space.
Under this assumption, clean 3D point clouds lie on a low-dimensional manifold, while adversarial examples are displaced from it.

As shown in Fig. \ref{manifold}, the method restores perturbed samples by projecting them back onto the data manifold using feature space similarity to the training set.
Since the manifold is not explicitly known, the method approximates it using features extracted from the training data and classifies each input by aggregating the softmax outputs of its nearest neighbors.

To enhance inference reliability, three weighting functions—uniform, entropy-based, and diversity-based—are employed to balance neighbor contributions based on prediction confidence.

Unlike many existing approaches, KNN-Defense does not rely on handcrafted geometric priors, architectural modifications, or retraining. Its lightweight and architecture-agnostic design enables fast inference and practical scalability. Empirical results confirm that KNN-Defense improves robustness against both point-shifting and point-dropping attacks across multiple architectures, establishing its effectiveness as a scalable solution for 3D point cloud classification.

\section{Related Work}\label{related}

In this section, an overview of 3D point clouds and commonly used distance metrics is provided. Previous research on 3D point cloud classification is summarized, and existing adversarial attack and defense methods for 3D point clouds are also discussed.

\subsection{Point Clouds and their Distance Metrics}

A point cloud is a data structure used to represent multi-dimensional data, including 3D objects.  Specifically, 3D point clouds consist of points that are sampled on an object surface. These points can be represented as vectors that contain the coordinates of the point along the $x$, $y$, and $z$ axes. . For example, a point cloud with $n$ points can be defined as $P = \{p_i | i = 1, 2, \ldots, n\}$, where $p_i$ represents the $i$ th point in the point cloud. In addition to the geometric coordinates, external information like color or intensity can also be saved in a point cloud.

Since point clouds are unordered sets of points, traditional distance metrics used for 2D images are not directly applicable unless a one-to-one mapping between the two point clouds is established. Two of the most popular methods for computing the distance between point clouds are the Hausdorff and Chamfer distances. Hausdorff distance measures the maximum distance of a point in one cloud to its nearest point in the other cloud, while the Chamfer distance uses the summation function instead of the maximum.

\subsection{Deep Learning Models for Point Cloud Classification}

PointNet \cite{qi2017pointnet} was the first deep network that was applied directly to 3D point clouds. It uses multilayer perceptrons and max-pooling (as a symmetric function) to extract global features of each point cloud. PointNet++ \cite{qi2017pointnet++} in contrast, exploits PointNet in a hierarchical structure to consider the local features in classification. Dynamic Graph CNN (DGCNN) \cite{wang2019dynamic} boosted the accuracy even further by constructing a graph of $k$ nearest-neighbors of each point and applying EdgeConv to these graphs, so that it can capture the geometric structure of point clouds as well as local features. Point Cloud Transformer (PCT) \cite{guo2021pct} adapts the transformer architecture to handle 3D point cloud data by transforming point clouds into a high-dimensional feature space. Using the self-attention mechanism, it weighs the influence of all other points (based on similarity) when processing a specific one. CurveNet \cite{xiang2021walk} is another architecture that achieves high accuracy by leveraging the aggregation of hypothetical curves within point clouds. Despite their architectural diversity and performance, all of these networks remain vulnerable to adversarial examples, through imperceptible changes that deceive the model.

\subsection{Adversarial Attacks Methods}

Based on the Carlini \& Wagner (C \& W) \cite{carlini2017towards} optimization-based framework, \cite{xiang2019generating} proposed various methods for generating adversarial point clouds through point shifting (using $l_2$ norm) and point adding (using Chamfer and Hausdorff distances) attacks. The KNN attack \cite{tsai2020robust} combines both KNN and Chamfer distance metrics to generate constructible adversarial examples. Additionally, well-known 2D gradient-based attacking algorithms such as FGSM \cite{goodfellow2015explaining}, JSMA \cite{papernot2016limitations}, and PGD \cite{madry2018towards} have been adapted for use against 3D classification networks \cite{liu2019extending}. \cite{zheng2019pointcloud} developed a point dropping attack by identifying the saliency map of point clouds and removing corresponding points. AdvPC \cite{hamdi2020advpc} and ShapeAdv \cite{lee2020shapeadv} utilized autoencoders to generate adversarial point clouds. LG-GAN \cite{zhou2020lg} introduced a novel algorithm based on Generative Adversarial Networks (GANs) for this purpose. Furthermore, AOF \cite{liu2022boosting} decomposes inputs into low- and high-frequency components, and generates adversarial point clouds by manipulating the low-frequency ones.

\subsection{Adversarial Defense Methods}

In the field of 3D adversarial defense, adversarial training \cite{liu2019extending} is utilized to train classification models using adversarial examples, aiming to enhance their robustness against various types of attacks. In \cite{yang2019adversarial}, an approach is presented where Gaussian noise is added to the adversarial point clouds to push them out of the adversarial sub-space. Another method proposed in the same study, known as SRS, involves randomly selecting and removing $n$ points from the point clouds. In contrast, the SOR (Statistical Outlier Removal) \cite{zhou2019dup} method focuses on selectively removing outlier points. DUP-Net \cite{zhou2019dup}, which operates based on the manifold assumption, incorporates the use of SOR as a preprocessing step to remove outliers. The remaining points are then fed into an upsampler network to obtain the final classification. \cite{wu2020if} introduces two methods to restore the clean point cloud, leveraging deep implicit functions. Following the SOR pre-processing step, the first method employs implicit function networks to construct a mesh from the input point clouds and subsequently samples a number of points from this mesh. The latter approach restores the original shape by using geometry-aware and distribution-aware loss functions. By applying a random transformation to the indices of the point clouds, \cite{zhang2023art} changes the gradients that an attacker would calculate, making adversarial examples ineffective while the classifier's ability to interpret the data remains unaffected. Moreover, \cite{zhang2023ada3diff} presents Ada3Diff, which restores clean data from adversarial point clouds using probabilistic diffusion models [13]. It first estimates each point’s perturbation to gauge distortion, then incrementally adds noise and reverses it to recover the original clean distribution.
\cite{zhang2022pointcutmix} introduces PointCutMix, which improves model robustness and generalization by mixing point clouds based on optimal correspondences computed via Earth Mover’s Distance (EMD).Lastly, \cite{naderi2023lpf} uses frequency analysis to boost robustness: it applies spherical harmonics to capture point-cloud frequencies, then a low-pass filter to isolate core shape components, and feeds these filtered clouds to a classifier to avoid adversarial high-frequency noise.

A related line of work in 2D adversarial defense is the Web-scale Nearest-neighbor Search proposed by Dubey et al. \cite{dubey2019defense}. The key concept behind this method is the manifold assumption. Inspired by this approach, the present work extends the nearest-neighbor principle to the 3D domain, adapting it to unordered and irregular point cloud structures to achieve robustness under 3D adversarial perturbations.

\section{Proposed Method}

This section presents the proposed 3D defense approach against adversarial examples in point cloud classification. The method operates under the manifold assumption, which posits that clean point clouds reside on a low-dimensional manifold, while adversarial examples are displaced away from this structure. The primary objective is to project these perturbed inputs back onto the data manifold by leveraging semantically similar neighbors from the training set. To improve this projection, three weighting strategies are explored.

\begin{figure}[t!]
\centering
\includegraphics[width=0.9\textwidth]{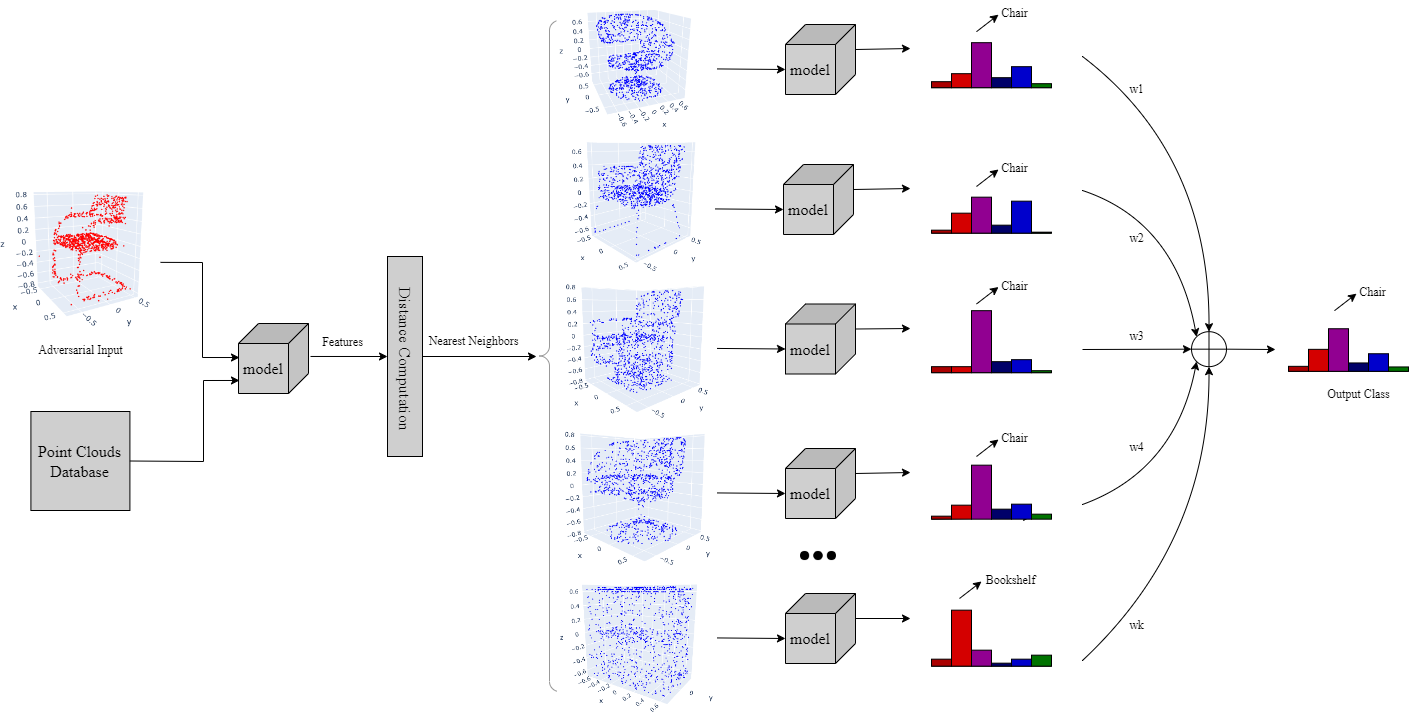}
\caption{Pipeline of proposed 3D KNN-Defense method. It involves finding the k nearest neighbors of a given input in the point cloud database based on the distance between the global feature vectors. The weighted average of the neighbors’ softmax vectors is used to predict the final class label.}
\label{architecture}
\end{figure}

\subsection{Defense Approach}

The core assumption behind the proposed defense method is that adversarial attacks cause the input point cloud to deviate from the natural point cloud manifold. Therefore, the objective is to identify the projection of each adversarial input onto the manifold and classify this projection instead of the adversarial input itself. However, the practical implementation of the approach approximates this projection, since the actual data manifold is unknown.

As illustrated in Fig. \ref{architecture}, the input (either adversarial or clean) is fed into the classification network and features are extracted from specific hidden layers. Pairwise distances are then computed between the input features and those of each sample in the point cloud database to determine the nearest neighbors. Inspired by \cite{dubey2019defense}, the method computes the weighted average of the neighbors’ softmax vectors using a weighting function. The final class prediction corresponds to the maximum element in the resulting weighted average vector.

\begin{algorithm}[t!]
\caption{Defense using Nearest-Neighbor Search}\label{algo}
\textbf{Input:} Point Cloud $\textbf{\emph{P}}$, outputs of feature layer $\mathcal{L}$ in the classification model $\mathcal{F}$, distance metric $\mathcal{D}$, nearest neighbor number $k$, weighting function $W$, training point clouds dataset $S_{train}$ \\
\textbf{Output:} Final predicted class $c$
\begin{algorithmic}[1]
\State Initialize the nearest-neighbors set $\mathcal{N}(\textbf{\emph{P}}) = \varnothing$
\State Compute $\mathcal{L}(\textbf{\emph{P}})$ and $\mathcal{L}(\textbf{\emph{P}}_i')$ for each $\textbf{\emph{P}}_i' \in S_{train}$
\State $d_i(\textbf{\emph{P}}) = \mathcal{D}(\textbf{\emph{P}}, \textbf{\emph{P}}_i')$
\State Sort the values $d_i(\textbf{\emph{P}})$ in ascending order, and append the first $k$ neighbors to $\mathcal{N}(\textbf{\emph{P}})$
\State $\forall \textbf{\emph{P}}_j' \in \mathcal{N}(\textbf{\emph{P}})$: Calculate the softmax vector $s_{\mathcal{F}}(\textbf{\emph{P}}_j')$
\State $s_{avg} = \sum_{j=1}^{k} W(\textbf{\emph{P}}_j')s_{\mathcal{F}}(\textbf{\emph{P}}_j')$
\State \Return Prediction class $c = argmax(s_{avg})$
\end{algorithmic}
\end{algorithm}

Algorithm \ref{algo} outlines the steps involved in our KNN-Defense approach.

\subsection{Weighting Functions}

Different weighting schemes could be taken into account to compute the weighted average of the nearest neighbor's softmax vectors. Similar to \cite{dubey2019defense}, three distance metrics are employed in the proposed KNN-Defense method: Uniform Weighting (UW), Entropy-based Weighting (EW), and Diversity-based Weighting (DW). 

\subsubsection{Uniform Weighting (UW)}

The uniform weighting strategy assigns equal importance to all of the $k$ nearest neighbors, regardless of their semantic proximity to the input. This approach is computationally simple and stable, making it suitable for fast or resource-constrained scenarios. However, in certain cases—particularly when some neighbors belong to semantically inconsistent classes—uniform weighting may reduce classification accuracy. 

As illustrated in Fig.~\ref{architecture}, treating all neighbors equally can introduce noise into the aggregated softmax output. Nonetheless, despite its limitations, UW remains an effective baseline and contributes to the overall robustness of the ensemble defense system proposed in this work.

\subsubsection{Entropy-based Weighting (EW)}

The Entropy-based Weighting function assigns weights to samples based on the difference between their softmax vector and a uniform softmax vector, in which all elements are equal. The entropy of a softmax vector $s$ is computed using the equation:

\begin{equation}
    H(s) = -\sum_{c=1}^{C} s_c \log s_c,
    \label{entropy-eq}
\end{equation}
where $C$ represents the number of classes in the dataset, and $s_c$ corresponds to the value of class $c$ in the softmax vector.

Since all the values of $s_c$ in the uniform softmax vector are equal to $\frac{1}{C}$, the entropy of this vector can be determined as $\log C$ using Formula \parref{entropy-eq}. Therefore, the weight assigned to a given softmax vector $s$ is defined as:

\begin{equation}
    w = \lvert \log C + \sum_{c=1}^{C} s_c \log s_c \rvert.
    \label{ew-eq}
\end{equation}

This weighting strategy is particularly effective in filtering out uncertain predictions, making it well-suited for scenarios where noisy or ambiguous neighbors may be present.

\subsubsection{Diversity-based Weighting (DW)}

The DW (Diversity-based Weighting) metric evaluates the weight of a softmax vector, denoted as $s$, based on the gap between its maximum value and the next top $M$ values. This means that a softmax vector with a larger gap between its maximum element and the other top $M$ elements will be assigned a higher weight. The DW metric employs the following function to determine the weight of a sorted softmax vector, $\hat{s}$, arranged in descending order:

\begin{equation}
    w = \sum_{m=2}^{M+1} (\hat{s}_1 - \hat{s}_m)^P.
    \label{dw-eq}
\end{equation}

In this equation, $P$ and $M$ are the two parameters, set to $P = 3$ and $M = 20$ in the experiments. By emphasizing softmax distributions with dominant class predictions, DW helps reinforce high-confidence decisions, which is especially valuable in the presence of class ambiguity or adversarial noise.

\section{Experiments} \label{experiments}
In this section, the effectiveness of the proposed model is evaluated. The results are compared with those of existing defenses to highlight the improvements that have been achieved. In addition, a detailed explanation of the experimental setup and an in-depth analysis of the performance and effectiveness of the proposed method in relation to existing approaches are provided.

\subsection{Dataset and Settings}

In the experiments, two datasets were leveraged: ModelNet40 \cite{wu20153d} and ScanObjectNN \cite{uy2019revisiting}. ModelNet40 consists of 12,311 CAD models of 40 different object categories. 9,843 of these objects were used for training, while the remaining 2,468 were kept for testing. The augmented version of this dataset, as provided in\cite{xiang2019generating}, was employed. In this version, 1,024 points are randomly sampled from each object’s surface and normalized to fit within a unit sphere. ScanObjectNN, on the other hand, contains objects in 15 classes with a training set of 11,416 and a test set of 2,882.

The victim classification models used in this study include PointNet \cite{qi2017pointnet}, PointNet++ \cite{qi2017pointnet++}, DGCNN \cite{wang2019dynamic}, and PCT \cite{guo2021pct}. For targeted attacks, Shift-L2 \cite{xiang2019generating}, Add-Chamfer \cite{xiang2019generating}, Add-Hausdorff \cite{xiang2019generating}, and Shift-KNN \cite{tsai2020robust} were employed. For untargeted attacks, the point-dropping method \cite{zheng2019pointcloud} was applied to clean point clouds to generate adversarial examples. Both targeted and untargeted versions of AdvPC \cite{hamdi2020advpc} and AOF \cite{liu2022boosting} were also included in the experiments. Among the existing 3D defense algorithms, SRS \cite{yang2019adversarial}, SOR \cite{zhou2019dup}, DUP-Net \cite{zhou2019dup}, and IF-Defense \cite{wu2020if} were evaluated for comparison with the proposed KNN-Defense method.

In ModelNet40 targeted attack scenarios, and following the procedure described in \cite{xiang2019generating}, 25 samples were randomly selected from each of the 10 largest classes in the ModelNet40 dataset. For each selected sample, attacks were performed considering the remaining 9 classes as target labels, resulting in the generation of 2,250 adversarial examples per targeted attack method. In contrast, untargeted attacks were applied to the entire test set.

\begin{figure}[t!]
\begin{center}
\subfloat[PointNet \cite{qi2017pointnet}]{\includegraphics[width=0.5\textwidth]{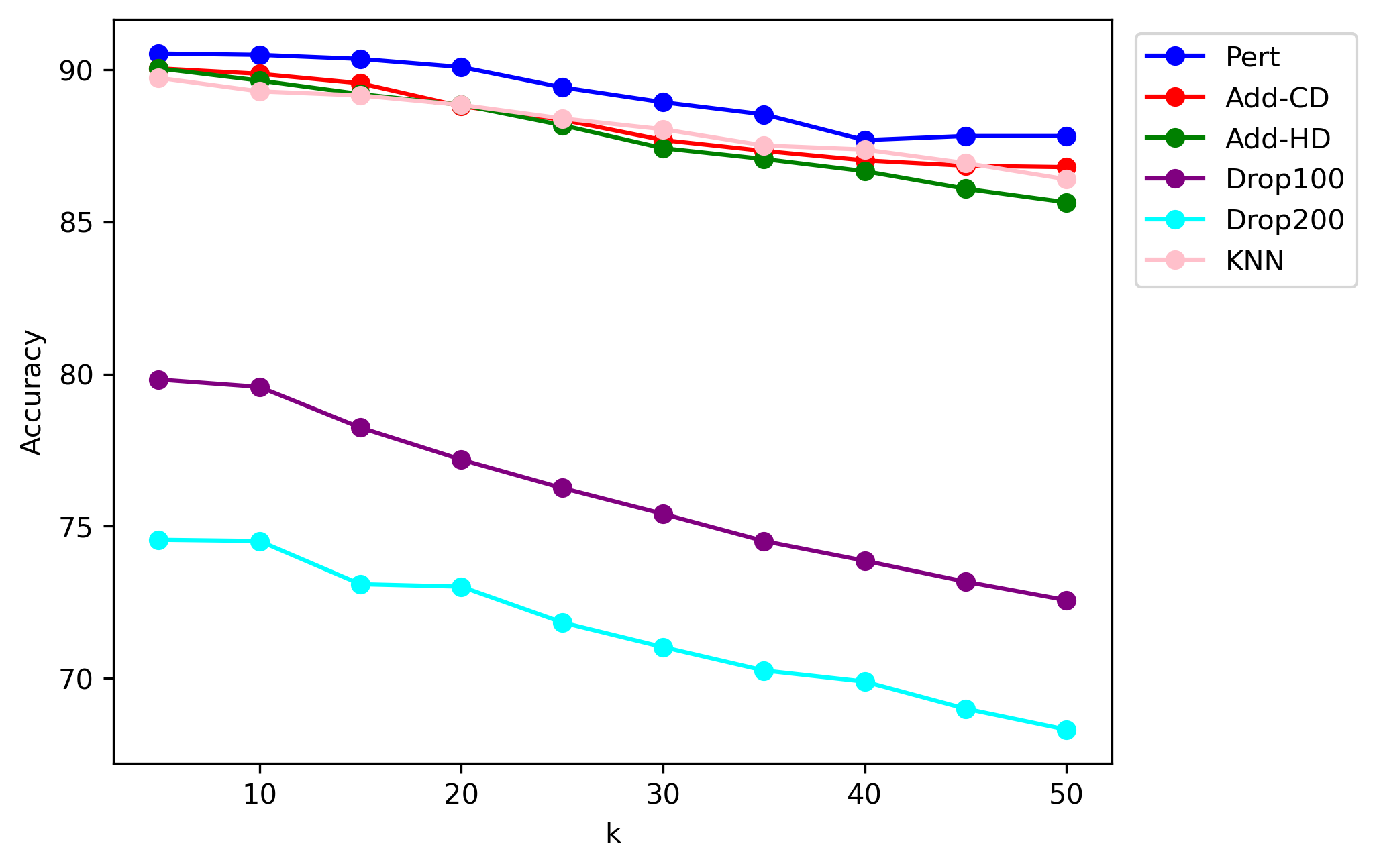}\label{pointnet-k-img}}%
\hfill
\subfloat[PointNet++ \cite{qi2017pointnet++}]{\includegraphics[width=0.5\textwidth]{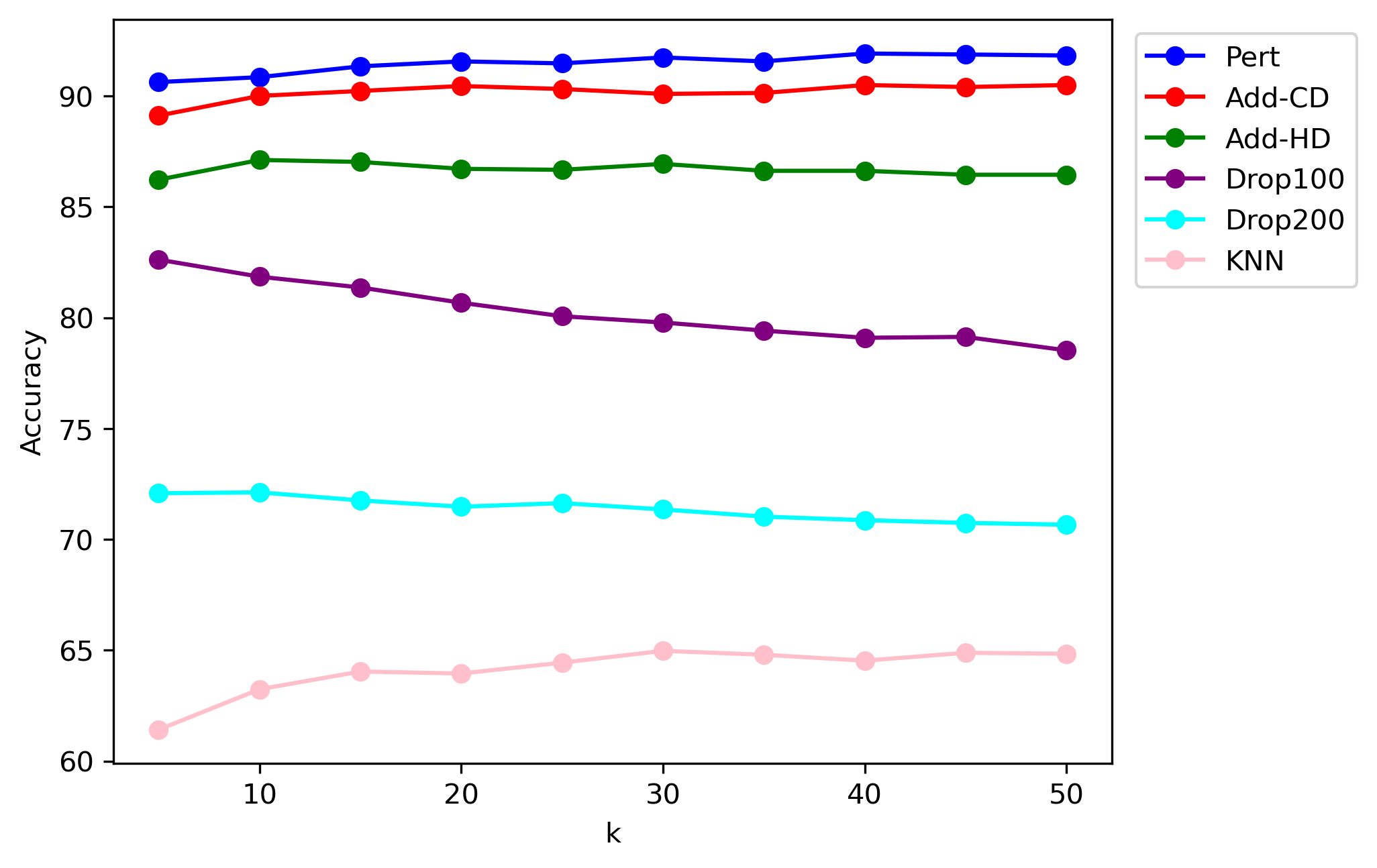}\label{pointnet2-k-img}}%
\\
\subfloat[DGCNN \cite{wang2019dynamic}]{\includegraphics[width=0.5\textwidth]{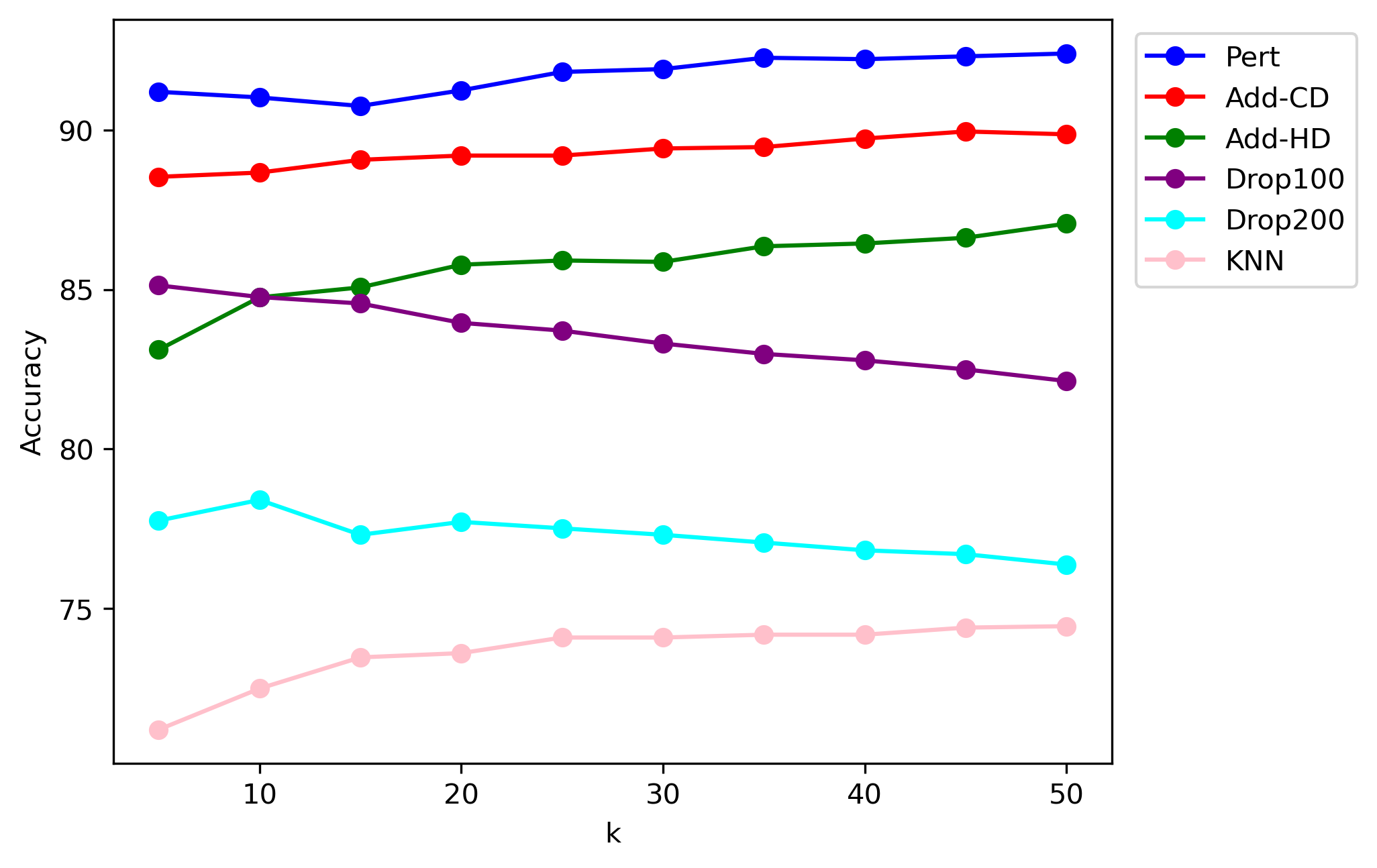}\label{dgcnn-k-img}}%
\hfill
\subfloat[PCT \cite{guo2021pct}]{\includegraphics[width=0.5\textwidth]{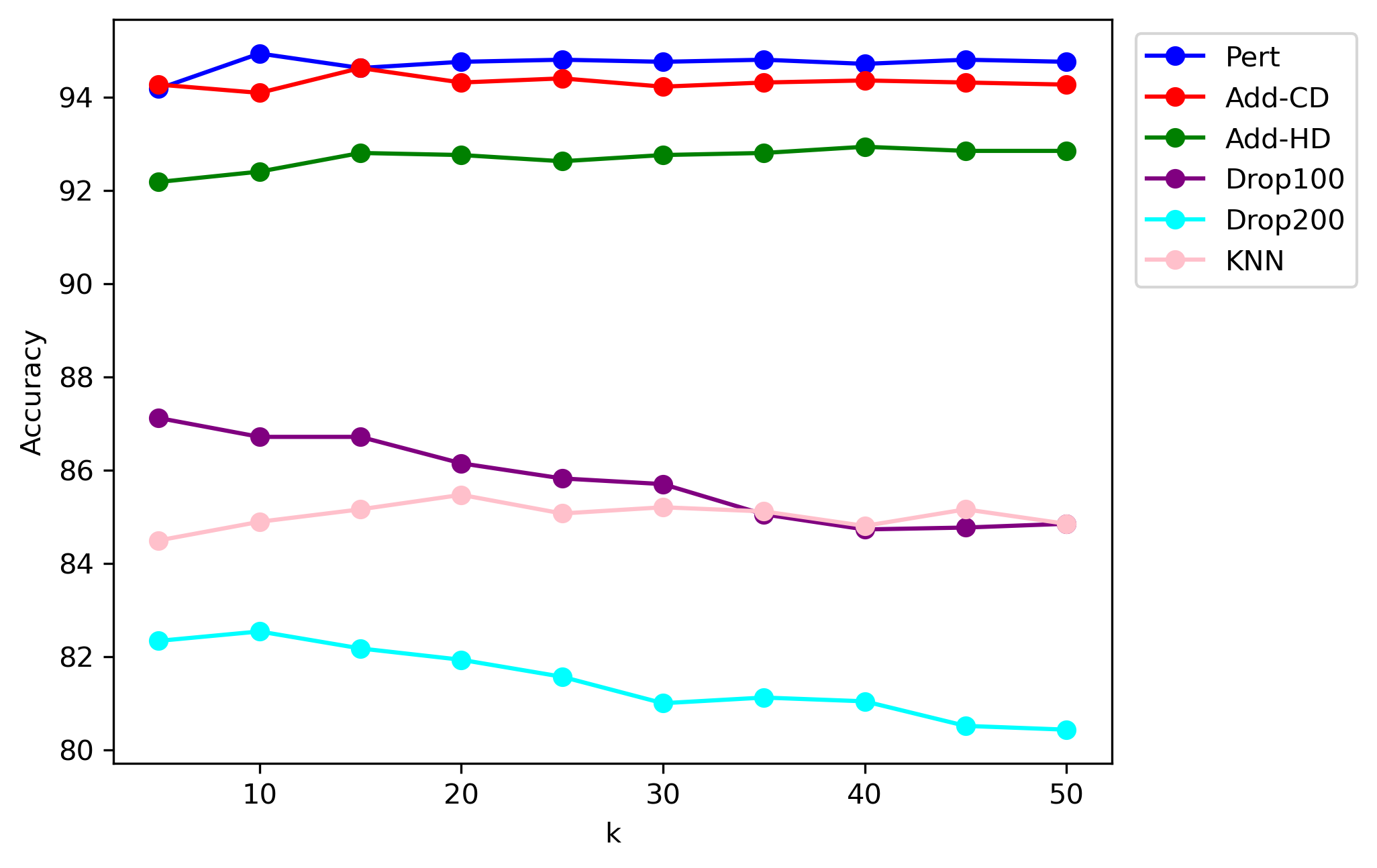}\label{pct-k-img}}%
\end{center}
\caption{Accuracy of different classification models against various attacks, considering different values of $k$, using the EW metric as the weighting function.}
\label{bestk}%
\end{figure}

In the KNN-Defense method, the outputs of feature aggregation layers are used as the learned representation of a point cloud sample. To determine the optimal $k$ parameter, several experiments were conducted. As shown in Fig. \ref{bestk}, and based on the EW metric, the values selected were $k = 5$ for PointNet \cite{qi2017pointnet}, $k = 10$ for PCT \cite{guo2021pct}, and $k = 15$ for both PointNet++ \cite{qi2017pointnet++} and DGCNN \cite{wang2019dynamic}.

\subsection{Comparison with State-of-the-art Defenses}

In this section, the classification accuracy of proposed KNN-Defense method is compared with SRS \cite{yang2019adversarial}, SOR \cite{zhou2019dup}, DUP-Net \cite{zhou2019dup}, and IF-Defense \cite{wu2020if}. 

The classification accuracy of different defense algorithms against targeted and untargeted attacks on PointNet \cite{qi2017pointnet}, PointNet++ \cite{qi2017pointnet++}, DGCNN \cite{wang2019dynamic}, and PCT \cite{guo2021pct} is provided in Table \ref{table-pointnet}, Table \ref{table-pointnet2}, Table \ref{table-dgcnn}, and Table \ref{table-pct} respectively. Moreover, the results of KNN-Defense are reported using each of the three weighting functions introduced earlier. The highest accuracy value is shown in bold, while the second highest is highlighted in blue.

\begin{table}[t!]
\captionsetup{width=\textwidth, justification=justified}
\caption{Accuracy comparison of different defense methods against adversarial attacks on PointNet \cite{qi2017pointnet} and ModelNet40 datasets.}\label{table-pointnet}
\setlength{\tabcolsep}{0.2\tabcolsep}
\begin{tabular}{|c|c|cccccc|cccc|}
\hline
            \multicolumn{2}{|c|}{}      & \multicolumn{6}{c|}{\textbf{Targeted Attacks}}                                                                                                                                                                                                                                                                                                                                                                      & \multicolumn{4}{c|}{\textbf{Untargeted attacks}}                                                                                                                                                                             \\ \hline
\textbf{Defenses} & \multicolumn{1}{c|}{Clean}                                 & \multicolumn{1}{c|}{Pert}                                  & \multicolumn{1}{c|}{KNN}                                   & \multicolumn{1}{c|}{Add-HD}                                & \multicolumn{1}{c|}{Add-CD}                                & \multicolumn{1}{c|}{AdvPC}                                 & AOF                                   & \multicolumn{1}{c|}{Drop100}                               & \multicolumn{1}{c|}{Drop200}                               & \multicolumn{1}{c|}{AdvPC}                                 & AOF                                   \\ \hline
No defense        & \multicolumn{1}{c|}{{\color[HTML]{333333} \textbf{88.09}}} & \multicolumn{1}{c|}{{\color[HTML]{333333} 0.00}}           & \multicolumn{1}{c|}{{\color[HTML]{333333} 0.89}}           & \multicolumn{1}{c|}{{\color[HTML]{333333} 0.00}}           & \multicolumn{1}{c|}{{\color[HTML]{333333} {0.00}}}  & \multicolumn{1}{c|}{{\color[HTML]{333333} 0.04}}           & {\color[HTML]{333333} 0.00}           & \multicolumn{1}{c|}{{\color[HTML]{333333} 57.09}}          & \multicolumn{1}{c|}{{\color[HTML]{333333} 27.11}}          & \multicolumn{1}{c|}{{\color[HTML]{333333} 1.34}}           & {\color[HTML]{333333} 0.00}           \\ \hline
SRS               & \multicolumn{1}{c|}{{\color[HTML]{333333} 76.05}}          & \multicolumn{1}{c|}{{\color[HTML]{333333} 76.27}}          & \multicolumn{1}{c|}{{\color[HTML]{333333} 27.87}}          & \multicolumn{1}{c|}{{\color[HTML]{333333} 75.42}}          & \multicolumn{1}{c|}{{\color[HTML]{333333} 74.36}}          & \multicolumn{1}{c|}{{\color[HTML]{333333} 47.33}}          & {\color[HTML]{333333} 9.38}           & \multicolumn{1}{c|}{{\color[HTML]{333333} 57.78}}          & \multicolumn{1}{c|}{{\color[HTML]{333333} 29.54}}          & \multicolumn{1}{c|}{{\color[HTML]{333333} 26.99}}          & {\color[HTML]{333333} 4.78}           \\ \hline
SOR               & \multicolumn{1}{c|}{{\color[HTML]{333333} 76.70}}          & \multicolumn{1}{c|}{{\color[HTML]{333333} 82.71}}          & \multicolumn{1}{c|}{{\color[HTML]{333333} 64.00}}          & \multicolumn{1}{c|}{{\color[HTML]{333333} 83.38}}          & \multicolumn{1}{c|}{{\color[HTML]{333333} 83.20}}          & \multicolumn{1}{c|}{{\color[HTML]{333333} 72.09}}          & {\color[HTML]{333333} 38.27}          & \multicolumn{1}{c|}{{\color[HTML]{333333} 58.14}}          & \multicolumn{1}{c|}{{\color[HTML]{333333} 31.52}}          & \multicolumn{1}{c|}{{\color[HTML]{333333} 48.87}}          & {\color[HTML]{333333} 17.18}          \\ \hline
DUP-Net           & \multicolumn{1}{c|}{{\color[HTML]{333333} 77.67}}          & \multicolumn{1}{c|}{{\color[HTML]{333333} 83.24}}          & \multicolumn{1}{c|}{{\color[HTML]{333333} 79.24}}          & \multicolumn{1}{c|}{{\color[HTML]{333333} 84.71}}          & \multicolumn{1}{c|}{{\color[HTML]{333333} 84.44}}          & \multicolumn{1}{c|}{{\color[HTML]{333333} 79.20}}          & {\color[HTML]{333333} 61.33}          & \multicolumn{1}{c|}{{\color[HTML]{333333} 63.29}}          & \multicolumn{1}{c|}{{\color[HTML]{333333} 40.15}}          & \multicolumn{1}{c|}{{\color[HTML]{333333} 63.86}}          & {\color[HTML]{333333} 45.70}          \\ \hline
IF-Defense        & \multicolumn{1}{c|}{{\color[HTML]{3531FF} 84.64}}          & \multicolumn{1}{c|}{{\color[HTML]{333333} 90.22}}          & \multicolumn{1}{c|}{{\color[HTML]{333333} 89.24}}          & \multicolumn{1}{c|}{{\color[HTML]{3531FF} 90.00}} & \multicolumn{1}{c|}{{\color[HTML]{333333} 89.42}}          & \multicolumn{1}{c|}{{\color[HTML]{3531FF} 88.76}} & {\color[HTML]{333333} 79.42}          & \multicolumn{1}{c|}{{\color[HTML]{333333} 71.88}}          & \multicolumn{1}{c|}{{\color[HTML]{333333} 54.54}}          & \multicolumn{1}{c|}{{\color[HTML]{333333} \textbf{82.94}}} & {\color[HTML]{333333} \textbf{72.73}} \\ \hline
Ours(UW)          & \multicolumn{1}{c|}{{\color[HTML]{333333} 83.55}}          & \multicolumn{1}{c|}{{\color[HTML]{333333} \textbf{90.71}}} & \multicolumn{1}{c|}{{\color[HTML]{333333} \textbf{89.78}}} & \multicolumn{1}{c|}{{\color[HTML]{333333} \textbf{90.04}}}          & \multicolumn{1}{c|}{{\color[HTML]{333333} \textbf{90.22}}} & \multicolumn{1}{c|}{{\color[HTML]{333333} 88.67}}          & {\color[HTML]{333333} \textbf{82.89}} & \multicolumn{1}{c|}{{\color[HTML]{3531FF} 79.78}}          & \multicolumn{1}{c|}{{\color[HTML]{333333} \textbf{74.64}}} & \multicolumn{1}{c|}{{\color[HTML]{333333} 81.36}}          & {\color[HTML]{333333} 70.26}          \\ \hline
Ours(EW)          & \multicolumn{1}{c|}{{\color[HTML]{333333} 83.55}}          & \multicolumn{1}{c|}{{\color[HTML]{333333} 90.53}}          & \multicolumn{1}{c|}{{\color[HTML]{3531FF} 89.73}}          & \multicolumn{1}{c|}{{\color[HTML]{333333} \textbf{90.04}}}          & \multicolumn{1}{c|}{{\color[HTML]{3531FF} 90.04}}          & \multicolumn{1}{c|}{{\color[HTML]{3531FF} 88.76}}          & {\color[HTML]{333333} 82.76}          & \multicolumn{1}{c|}{{\color[HTML]{333333} \textbf{79.82}}} & \multicolumn{1}{c|}{{\color[HTML]{3531FF} 74.55}}          & \multicolumn{1}{c|}{{\color[HTML]{3531FF} 81.40}}          & {\color[HTML]{3531FF} 70.46}          \\ \hline
Ours(DW)          & \multicolumn{1}{c|}{{\color[HTML]{333333} 83.55}}          & \multicolumn{1}{c|}{{\color[HTML]{3531FF} 90.62}}          & \multicolumn{1}{c|}{{\color[HTML]{333333} 89.69}}          & \multicolumn{1}{c|}{{\color[HTML]{333333} 89.96}}          & \multicolumn{1}{c|}{{\color[HTML]{333333} 89.82}}          & \multicolumn{1}{c|}{{\color[HTML]{333333} \textbf{88.80}}}          & {\color[HTML]{3531FF} 82.80}          & \multicolumn{1}{c|}{{\color[HTML]{333333} \textbf{79.82}}} & \multicolumn{1}{c|}{{\color[HTML]{333333} 74.51}}          & \multicolumn{1}{c|}{{\color[HTML]{333333} 81.20}}          & {\color[HTML]{333333} 70.38}          \\ \hline
\end{tabular}
\end{table}

\begin{table}[t!]
\captionsetup{width=\textwidth, justification=justified}
\caption{Accuracy comparison of different defense methods against adversarial attacks on PointNet++ \cite{qi2017pointnet++} and ModelNet40 datasets.}\label{table-pointnet2}
\setlength{\tabcolsep}{0.2\tabcolsep}
\begin{tabular}{|c|c|cccccc|cccc|}
\hline
            \multicolumn{2}{|c|}{}      & \multicolumn{6}{c|}{\textbf{Targeted Attacks}}                                                                                                                                                                                                                                                                                                                                                                      & \multicolumn{4}{c|}{\textbf{Untargeted attacks}}                                                                                                                                                                             \\ \hline
\textbf{Defenses} & \multicolumn{1}{c|}{Clean}                                 & \multicolumn{1}{c|}{Pert}                                  & \multicolumn{1}{c|}{KNN}                                   & \multicolumn{1}{c|}{Add-HD}                                & \multicolumn{1}{c|}{Add-CD}                                & \multicolumn{1}{c|}{AdvPC}                                 & AOF                                   & \multicolumn{1}{c|}{Drop100}                               & \multicolumn{1}{c|}{Drop200}                               & \multicolumn{1}{c|}{AdvPC}                                 & AOF                                   \\ \hline
No defense        & \multicolumn{1}{c|}{{\color[HTML]{333333} \textbf{89.18}}} & \multicolumn{1}{c|}{{\color[HTML]{333333} 51.20}}          & \multicolumn{1}{c|}{{\color[HTML]{333333} 0.00}}           & \multicolumn{1}{c|}{{\color[HTML]{333333} 41.91}}          & \multicolumn{1}{c|}{{\color[HTML]{333333} 64.62}}          & \multicolumn{1}{c|}{{\color[HTML]{333333} 10.76}}          & {\color[HTML]{333333} 3.78}           & \multicolumn{1}{c|}{{\color[HTML]{333333} 77.55}}          & \multicolumn{1}{c|}{{\color[HTML]{333333} 62.72}}          & \multicolumn{1}{c|}{{\color[HTML]{333333} 16.21}}          & {\color[HTML]{333333} 1.50}           \\ \hline
SRS               & \multicolumn{1}{c|}{{\color[HTML]{333333} 81.65}}          & \multicolumn{1}{c|}{{\color[HTML]{333333} 84.40}}          & \multicolumn{1}{c|}{{\color[HTML]{333333} 61.11}}          & \multicolumn{1}{c|}{{\color[HTML]{333333} 78.09}}          & \multicolumn{1}{c|}{{\color[HTML]{333333} 85.82}}          & \multicolumn{1}{c|}{{\color[HTML]{333333} 50.18}}          & {\color[HTML]{333333} 23.78}          & \multicolumn{1}{c|}{{\color[HTML]{333333} 64.91}}          & \multicolumn{1}{c|}{{\color[HTML]{333333} 42.10}}          & \multicolumn{1}{c|}{{\color[HTML]{333333} 50.61}}          & {\color[HTML]{333333} 25.45}          \\ \hline
SOR               & \multicolumn{1}{c|}{{\color[HTML]{333333} 83.79}}          & \multicolumn{1}{c|}{{\color[HTML]{333333} 86.31}}          & \multicolumn{1}{c|}{{\color[HTML]{333333} 48.98}}          & \multicolumn{1}{c|}{{\color[HTML]{333333} \textbf{88.62}}} & \multicolumn{1}{c|}{{\color[HTML]{333333} 89.64}}          & \multicolumn{1}{c|}{{\color[HTML]{333333} 61.60}}          & {\color[HTML]{333333} 34.40}          & \multicolumn{1}{c|}{{\color[HTML]{333333} 73.70}}          & \multicolumn{1}{c|}{{\color[HTML]{333333} 62.80}}          & \multicolumn{1}{c|}{{\color[HTML]{333333} 52.76}}          & {\color[HTML]{333333} 19.61}          \\ \hline
DUP-Net           & \multicolumn{1}{c|}{{\color[HTML]{333333} 80.96}}          & \multicolumn{1}{c|}{{\color[HTML]{333333} 85.82}}          & \multicolumn{1}{c|}{{\color[HTML]{3531FF} 84.62}}          & \multicolumn{1}{c|}{{\color[HTML]{333333} 85.16}}          & \multicolumn{1}{c|}{{\color[HTML]{333333} 87.60}}          & \multicolumn{1}{c|}{{\color[HTML]{3531FF} 70.84}}          & {\color[HTML]{3531FF} 48.89}          & \multicolumn{1}{c|}{{\color[HTML]{333333} 72.81}}          & \multicolumn{1}{c|}{{\color[HTML]{333333} 64.79}}          & \multicolumn{1}{c|}{{\color[HTML]{333333} 59.68}}          & {\color[HTML]{3531FF} 37.28}          \\ \hline
IF-Defense        & \multicolumn{1}{c|}{{\color[HTML]{333333} 82.46}}          & \multicolumn{1}{c|}{{\color[HTML]{333333} 89.24}}          & \multicolumn{1}{c|}{{\color[HTML]{333333} \textbf{89.16}}} & \multicolumn{1}{c|}{{\color[HTML]{3531FF} 88.49}}          & \multicolumn{1}{c|}{{\color[HTML]{333333} 89.24}}          & \multicolumn{1}{c|}{{\color[HTML]{333333} \textbf{81.42}}} & {\color[HTML]{333333} \textbf{69.51}} & \multicolumn{1}{c|}{{\color[HTML]{333333} 77.19}}          & \multicolumn{1}{c|}{{\color[HTML]{333333} 68.52}}          & \multicolumn{1}{c|}{{\color[HTML]{333333} \textbf{75.53}}} & {\color[HTML]{333333} \textbf{62.68}} \\ \hline
Ours(UW)          & \multicolumn{1}{c|}{{\color[HTML]{333333} 87.03}}          & \multicolumn{1}{c|}{{\color[HTML]{3531FF} 91.29}}          & \multicolumn{1}{c|}{{\color[HTML]{333333} 63.69}}          & \multicolumn{1}{c|}{{\color[HTML]{333333} 86.93}}          & \multicolumn{1}{c|}{{\color[HTML]{333333} \textbf{90.58}}} & \multicolumn{1}{c|}{{\color[HTML]{333333} 68.00}}          & {\color[HTML]{333333} 39.78}          & \multicolumn{1}{c|}{{\color[HTML]{333333} 81.00}}          & \multicolumn{1}{c|}{{\color[HTML]{333333} \textbf{72.12}}} & \multicolumn{1}{c|}{{\color[HTML]{333333} 60.66}}          & {\color[HTML]{333333} 34.44}          \\ \hline
Ours(EW)          & \multicolumn{1}{c|}{{\color[HTML]{333333} 87.16}}          & \multicolumn{1}{c|}{{\color[HTML]{333333} \textbf{91.33}}} & \multicolumn{1}{c|}{{\color[HTML]{333333} 63.51}}          & \multicolumn{1}{c|}{{\color[HTML]{333333} 86.84}}          & \multicolumn{1}{c|}{{\color[HTML]{3531FF} 90.44}}          & \multicolumn{1}{c|}{{\color[HTML]{333333} 67.96}}          & {\color[HTML]{333333} 40.04}          & \multicolumn{1}{c|}{{\color[HTML]{3531FF} 81.08}}          & \multicolumn{1}{c|}{{\color[HTML]{333333} 72.04}}          & \multicolumn{1}{c|}{{\color[HTML]{333333} 60.78}}          & {\color[HTML]{333333} 34.60}          \\ \hline
Ours(DW)          & \multicolumn{1}{c|}{{\color[HTML]{3531FF} 87.24}}          & \multicolumn{1}{c|}{{\color[HTML]{333333} 91.16}}          & \multicolumn{1}{c|}{{\color[HTML]{333333} 63.20}}          & \multicolumn{1}{c|}{{\color[HTML]{333333} 86.98}}          & \multicolumn{1}{c|}{{\color[HTML]{333333} 90.27}}          & \multicolumn{1}{c|}{{\color[HTML]{333333} 68.31}}          & {\color[HTML]{333333} 41.07}          & \multicolumn{1}{c|}{{\color[HTML]{333333} \textbf{81.20}}} & \multicolumn{1}{c|}{{\color[HTML]{3531FF} 72.08}}          & \multicolumn{1}{c|}{{\color[HTML]{3531FF} 60.98}}          & {\color[HTML]{333333} 34.48}          \\ \hline
\end{tabular}
\end{table}

\begin{table}[t!]
\captionsetup{width=\textwidth, justification=justified}
\caption{Accuracy comparison of different defense methods against adversarial attacks on DGCNN \cite{wang2019dynamic} and ModelNet40 datasets.}\label{table-dgcnn}
\setlength{\tabcolsep}{0.2\tabcolsep}
\begin{tabular}{|c|c|cccccc|cccc|}
\hline
          \multicolumn{2}{|c|}{}        & \multicolumn{6}{c|}{\textbf{Targeted Attacks}}                                                                                                                                                                                                                                                                                                       & \multicolumn{4}{c|}{\textbf{Untargeted attacks}}                                                                                                                                         \\ \hline
\textbf{Defenses} & \multicolumn{1}{c|}{Clean}                        & \multicolumn{1}{c|}{Pert}                     & \multicolumn{1}{c|}{KNN}                          & \multicolumn{1}{c|}{Add-HD}                       & \multicolumn{1}{c|}{Add-CD}                       & \multicolumn{1}{c|}{AdvPC}                        & \multicolumn{1}{c|}{AOF}     & \multicolumn{1}{c|}{Drop100}                     & \multicolumn{1}{c|}{Drop200}                     & \multicolumn{1}{c|}{AdvPC}                        & \multicolumn{1}{c|}{AOF}     \\ \hline
No defense        & \multicolumn{1}{c|}{\textbf{91.86}}               & \multicolumn{1}{c|}{0.00}                         & \multicolumn{1}{c|}{0.00}                         & \multicolumn{1}{c|}{0.00}                         & \multicolumn{1}{c|}{0.00}               & \multicolumn{1}{c|}{20.06}                        & 0.00                         & \multicolumn{1}{c|}{79.13}                        & \multicolumn{1}{c|}{63.01}                        & \multicolumn{1}{c|}{0.00}                         & 0.00                         \\ \hline
SRS               & \multicolumn{1}{c|}{87.07}                        & \multicolumn{1}{c|}{{\color[HTML]{3531FF} 91.33}} & \multicolumn{1}{c|}{{\color[HTML]{3531FF} 73.47}} & \multicolumn{1}{c|}{77.56}                        & \multicolumn{1}{c|}{{\color[HTML]{3531FF} 90.67}} & \multicolumn{1}{c|}{61.10}                        & 32.98                        & \multicolumn{1}{c|}{72.37}                        & \multicolumn{1}{c|}{56.73}                        & \multicolumn{1}{c|}{58.93}                        & 38.13                        \\ \hline
SOR               & \multicolumn{1}{c|}{88.49}                        & \multicolumn{1}{c|}{90.53}                        & \multicolumn{1}{c|}{17.29}                        & \multicolumn{1}{c|}{82.22} & \multicolumn{1}{c|}{82.98}                        & \multicolumn{1}{c|}{52.03}                        & 21.19                        & \multicolumn{1}{c|}{78.77}                        & \multicolumn{1}{c|}{68.11}                        & \multicolumn{1}{c|}{51.69}                        & 29.73                        \\ \hline
DUP-Net           & \multicolumn{1}{c|}{53.81}                        & \multicolumn{1}{c|}{50.80}                        & \multicolumn{1}{l|}{19.16} & \multicolumn{1}{c|}{47.16}                        & \multicolumn{1}{c|}{53.56}                        & \multicolumn{1}{c|}{30.51}                        & 19.85                        & \multicolumn{1}{c|}{44.08}                        & \multicolumn{1}{c|}{35.78}                        & \multicolumn{1}{c|}{23.73}                        & 13.91                        \\ \hline
IF-Defense        & \multicolumn{1}{c|}{87.32}                        & \multicolumn{1}{c|}{\textbf{93.51}}               & \multicolumn{1}{c|}{\textbf{90.00}}               & \multicolumn{1}{c|}{\textbf{90.36}}               & \multicolumn{1}{c|}{\textbf{92.40}}               & \multicolumn{1}{c|}{\textbf{80.59}}               & \textbf{68.23}               & \multicolumn{1}{c|}{81.97}                        & \multicolumn{1}{c|}{73.99}                        & \multicolumn{1}{c|}{\textbf{83.96}}               & \textbf{71.82}               \\ \hline
Ours(UW)          & \multicolumn{1}{c|}{88.98}                        & \multicolumn{1}{c|}{ 90.84} & \multicolumn{1}{c|}{73.38}                        & \multicolumn{1}{c|}{{\color[HTML]{3531FF} 85.07}} & \multicolumn{1}{c|}{89.07}                        & \multicolumn{1}{c|}{76.13}                        & 56.56                        & \multicolumn{1}{c|}{\textbf{84.64}}               & \multicolumn{1}{c|}{{\color[HTML]{3531FF} 77.43}} & \multicolumn{1}{c|}{81.33}                        & 68.58                        \\ \hline
Ours(EW)          & \multicolumn{1}{c|}{89.06}                        & \multicolumn{1}{c|}{90.76}                        & \multicolumn{1}{c|}{{\color[HTML]{3531FF} 73.47}} & \multicolumn{1}{c|}{{\color[HTML]{3531FF} 85.07}} & \multicolumn{1}{c|}{89.07} & \multicolumn{1}{c|}{{\color[HTML]{3531FF} 76.22}} & 56.73                        & \multicolumn{1}{c|}{{\color[HTML]{3531FF} 84.56}} & \multicolumn{1}{c|}{ 77.31} & \multicolumn{1}{c|}{81.42}                        & 68.84                        \\ \hline
Ours(DW)          & \multicolumn{1}{c|}{{\color[HTML]{3531FF} 89.10}} & \multicolumn{1}{c|}{90.84}                        & \multicolumn{1}{c|}{{\color[HTML]{3531FF} 73.47}} & \multicolumn{1}{c|}{{\color[HTML]{3531FF} 85.07}} & \multicolumn{1}{c|}{89.02}                        & \multicolumn{1}{c|}{76.18} & {\color[HTML]{3531FF} 56.93} & \multicolumn{1}{c|}{\textbf{84.64}}               & \multicolumn{1}{c|}{\textbf{77.47}}               & \multicolumn{1}{c|}{{\color[HTML]{3531FF} 81.82}} & {\color[HTML]{3531FF} 69.47} \\ \hline
\end{tabular}
\end{table}

\begin{table}[t!]
\captionsetup{width=\textwidth, justification=justified}
\caption{Accuracy comparison of different defense methods against adversarial attacks on PCT \cite{guo2021pct} and ModelNet40 datasets.}\label{table-pct}
\setlength{\tabcolsep}{0.2\tabcolsep}
\begin{tabular}{|c|c|cccccc|cccc|}
\hline
        \multicolumn{2}{|c|}{}          & \multicolumn{6}{c|}{\textbf{Targeted Attacks}}                                                                                                                                                                                                                                                                                                                                                                      & \multicolumn{4}{c|}{\textbf{Untargeted attacks}}                                                                                                                                                                             \\ \hline
\textbf{Defenses} & \multicolumn{1}{c|}{Clean}                                 & \multicolumn{1}{c|}{Pert}                                  & \multicolumn{1}{c|}{KNN}                                   & \multicolumn{1}{c|}{Add-HD}                                & \multicolumn{1}{c|}{Add-CD}                                & \multicolumn{1}{c|}{AdvPC}                                 & AOF                                   & \multicolumn{1}{c|}{Drop100}                               & \multicolumn{1}{c|}{Drop200}                               & \multicolumn{1}{c|}{AdvPC}                                 & AOF                                   \\ \hline
No defense        & \multicolumn{1}{c|}{{\color[HTML]{333333} \textbf{92.42}}} & \multicolumn{1}{c|}{{\color[HTML]{333333} 72.27}}          & \multicolumn{1}{c|}{{\color[HTML]{333333} 0.00}}           & \multicolumn{1}{c|}{{\color[HTML]{333333} 34.18}}          & \multicolumn{1}{c|}{{\color[HTML]{333333} 69.73}} & \multicolumn{1}{c|}{{\color[HTML]{333333} 8.49}}           & {\color[HTML]{333333} 2.80}           & \multicolumn{1}{c|}{{\color[HTML]{333333} 80.31}}          & \multicolumn{1}{c|}{{\color[HTML]{333333} 66.00}}          & \multicolumn{1}{c|}{{\color[HTML]{333333} 8.87}}           & {\color[HTML]{333333} 2.11}           \\ \hline
SRS               & \multicolumn{1}{c|}{{\color[HTML]{3531FF} 91.73}}          & \multicolumn{1}{c|}{{\color[HTML]{333333} 92.18}}          & \multicolumn{1}{c|}{{\color[HTML]{333333} 61.02}}          & \multicolumn{1}{c|}{{\color[HTML]{333333} 82.31}}          & \multicolumn{1}{c|}{{\color[HTML]{333333} 91.38}}          & \multicolumn{1}{c|}{{\color[HTML]{333333} 65.24}}          & {\color[HTML]{333333} 33.87}          & \multicolumn{1}{c|}{{\color[HTML]{333333} 83.23}}          & \multicolumn{1}{c|}{{\color[HTML]{333333} 74.59}}          & \multicolumn{1}{c|}{{\color[HTML]{333333} 60.17}}          & {\color[HTML]{333333} 21.96}          \\ \hline
SOR               & \multicolumn{1}{c|}{{\color[HTML]{333333} 91.65}}          & \multicolumn{1}{c|}{{\color[HTML]{333333} 93.73}}          & \multicolumn{1}{c|}{{\color[HTML]{333333} 25.02}}          & \multicolumn{1}{c|}{{\color[HTML]{333333} 90.49}}          & \multicolumn{1}{c|}{{\color[HTML]{333333} 93.69}}          & \multicolumn{1}{c|}{{\color[HTML]{333333} 65.6}}           & {\color[HTML]{333333} 33.96}          & \multicolumn{1}{c|}{{\color[HTML]{333333} 82.25}}          & \multicolumn{1}{c|}{{\color[HTML]{333333} 70.95}}          & \multicolumn{1}{c|}{{\color[HTML]{333333} 58.75}}          & {\color[HTML]{333333} 19.00}          \\ \hline
DUP-Net           & \multicolumn{1}{c|}{{\color[HTML]{333333} 87.44}}          & \multicolumn{1}{c|}{{\color[HTML]{333333} 92.98}}          & \multicolumn{1}{c|}{{\color[HTML]{333333} 79.11}}          & \multicolumn{1}{c|}{{\color[HTML]{333333} 87.16}}          & \multicolumn{1}{c|}{{\color[HTML]{333333} 92.44}}          & \multicolumn{1}{c|}{{\color[HTML]{333333} 61.11}}          & {\color[HTML]{333333} 38.22}          & \multicolumn{1}{c|}{{\color[HTML]{333333} 76.78}}          & \multicolumn{1}{c|}{{\color[HTML]{333333} 63.94}}          & \multicolumn{1}{c|}{{\color[HTML]{333333} 61.79}}          & {\color[HTML]{333333} 38.45}          \\ \hline
IF-Defense        & \multicolumn{1}{c|}{{\color[HTML]{333333} 88.65}}          & \multicolumn{1}{c|}{{\color[HTML]{333333} {92.71}}} & \multicolumn{1}{c|}{{\color[HTML]{333333} \textbf{90.89}}} & \multicolumn{1}{c|}{{\color[HTML]{333333} {91.38}}} & \multicolumn{1}{c|}{{\color[HTML]{333333} {92.00}}} & \multicolumn{1}{c|}{{\color[HTML]{333333} {87.07}}} & {\color[HTML]{333333} {75.60}}  & \multicolumn{1}{c|}{{\color[HTML]{3531FF} 83.59}}          & \multicolumn{1}{c|}{{\color[HTML]{333333} 75.49}}          & \multicolumn{1}{c|}{{\color[HTML]{333333} \textbf{82.09}}} & {\color[HTML]{333333} \textbf{68.11}} \\ \hline
Ours(UW)          & \multicolumn{1}{c|}{{\color[HTML]{333333} 90.52}}          & \multicolumn{1}{c|}{{\color[HTML]{3531FF} 94.76}}          & \multicolumn{1}{c|}{{\color[HTML]{3531FF} 84.89}}          & \multicolumn{1}{c|}{{\color[HTML]{333333} \textbf{92.58}}} & \multicolumn{1}{c|}{{\color[HTML]{3531FF} 94.36}}          & \multicolumn{1}{c|}{{\color[HTML]{333333} 88.67}}          & {\color[HTML]{333333} 81.29}          & \multicolumn{1}{c|}{{\color[HTML]{333333} \textbf{86.75}}} & \multicolumn{1}{c|}{{\color[HTML]{3531FF} 83.23}}          & \multicolumn{1}{c|}{{\color[HTML]{333333} 80.15}}          & {\color[HTML]{333333} 59.85}          \\ \hline
Ours(EW)          & \multicolumn{1}{c|}{{\color[HTML]{333333} 90.48}}          & \multicolumn{1}{c|}{{\color[HTML]{333333} \textbf{94.93}}} & \multicolumn{1}{c|}{{\color[HTML]{333333} 84.80}}          & \multicolumn{1}{c|}{{\color[HTML]{3531FF} 92.53}}          & \multicolumn{1}{c|}{{\color[HTML]{3531FF} 94.36}}          & \multicolumn{1}{c|}{{\color[HTML]{3531FF} 88.76}}          & {\color[HTML]{3531FF} 81.38}          & \multicolumn{1}{c|}{{\color[HTML]{333333} \textbf{86.75}}} & \multicolumn{1}{c|}{{\color[HTML]{333333} \textbf{83.27}}} & \multicolumn{1}{c|}{{\color[HTML]{333333} 80.19}}          & {\color[HTML]{333333} 59.76}          \\ \hline
Ours(DW)          & \multicolumn{1}{c|}{{\color[HTML]{333333} 90.76}}          & \multicolumn{1}{c|}{{\color[HTML]{333333} \textbf{94.93}}} & \multicolumn{1}{c|}{{\color[HTML]{333333} 84.80}}          & \multicolumn{1}{c|}{{\color[HTML]{333333} 92.76}}          & \multicolumn{1}{c|}{{\color[HTML]{333333} \textbf{94.40}}} & \multicolumn{1}{c|}{{\color[HTML]{333333} \textbf{88.89}}} & {\color[HTML]{333333} \textbf{82.31}} & \multicolumn{1}{c|}{{\color[HTML]{333333} \textbf{86.75}}} & \multicolumn{1}{c|}{{\color[HTML]{333333} {83.10}}} & \multicolumn{1}{c|}{{\color[HTML]{3531FF} 80.39}}          & {\color[HTML]{3531FF} 60.13}          \\ \hline
\end{tabular}
\end{table}

\begin{table}[t!]
\captionsetup{width=\textwidth, justification=justified}
\caption{Accuracy comparison of different defense methods against adversarial attacks on PCT \cite{guo2021pct} over ScanObjectNN \cite{uy2019revisiting} dataset.}\label{table-pct-scan}
\setlength{\tabcolsep}{0.2\tabcolsep}
\begin{tabular}{|c|c|cccc|cc|}
\hline
            &      & \multicolumn{4}{c|}{\textbf{Targeted Attacks}}                                                                                                                                                                                                                                                                                                                                                                      & \multicolumn{2}{c|}{\textbf{Untargeted attacks}}                                                                                                                                                                             \\ \hline
\textbf{Defenses} & Clean                                 & \multicolumn{1}{c|}{Pert}                                  & \multicolumn{1}{c|}{KNN}                                   & \multicolumn{1}{c|}{Add-HD}                                & Add-CD                                                  & \multicolumn{1}{c|}{Drop100}                               & Drop200                               \\ \hline
No defense        &  {\color[HTML]{333333} \textbf{76.66}} & \multicolumn{1}{c|}{{\color[HTML]{333333} 0.00}}          & \multicolumn{1}{c|}{{\color[HTML]{333333} 0.00}}           & \multicolumn{1}{c|}{{\color[HTML]{333333} 0.00}}                & {\color[HTML]{333333} 0.00}           & \multicolumn{1}{c|}{{\color[HTML]{333333} 63.87}}          & {\color[HTML]{333333} 53.30}           \\ \hline
SRS               & \multicolumn{1}{c|}{{\color[HTML]{3531FF} 73.66}}          & \multicolumn{1}{c|}{{\color[HTML]{333333} \textbf{68.32}}}          & \multicolumn{1}{c|}{{\color[HTML]{333333} 26.41}}          & \multicolumn{1}{c|}{{\color[HTML]{333333} 44.59}}          & {\color[HTML]{3531FF} 67.14}          & \multicolumn{1}{c|}{{\color[HTML]{333333} 66.20}}          & {\color[HTML]{333333} 60.72}          \\ \hline
SOR               & {\color[HTML]{333333} 73.49}          & \multicolumn{1}{c|}{{\color[HTML]{3531FF} 67.49}}          & \multicolumn{1}{c|}{{\color[HTML]{333333} 5.34}}          & \multicolumn{1}{c|}{{\color[HTML]{333333} \textbf{55.90}}}          & {\color[HTML]{333333} \textbf{68.84}}          & \multicolumn{1}{c|}{{\color[HTML]{333333} 65.58}}         & {\color[HTML]{333333} 58.40}          \\ \hline
DUP-Net           & {\color[HTML]{333333} 53.78}          & \multicolumn{1}{c|}{{\color[HTML]{333333} 46.53}}          & \multicolumn{1}{c|}{{\color[HTML]{333333} 38.76}}          & \multicolumn{1}{c|}{{\color[HTML]{333333} 42.16}}                  & {\color[HTML]{333333} 48.89}          & \multicolumn{1}{c|}{{\color[HTML]{333333} 48.72}}          & {\color[HTML]{333333} 44.27}          \\ \hline
IF-Defense        & {\color[HTML]{333333} 59.85}          & \multicolumn{1}{c|}{{\color[HTML]{333333} {57.18}}} & \multicolumn{1}{c|}{{\color[HTML]{333333} \textbf{53.26}}} & \multicolumn{1}{c|}{{\color[HTML]{3531FF} {54.82}}} & {\color[HTML]{333333} {58.19}}  & \multicolumn{1}{c|}{{\color[HTML]{333333} 55.03}}          & {\color[HTML]{333333} 50.07}           \\ \hline
Ours(UW)          & {\color[HTML]{333333} 70.44}         & \multicolumn{1}{c|}{{\color[HTML]{333333} 61.00}}          & \multicolumn{1}{c|}{{\color[HTML]{333333} 46.53}}          & \multicolumn{1}{c|}{{\color[HTML]{333333} 44.14}} & {\color[HTML]{333333} 61.07}          & \multicolumn{1}{c|}{{\color[HTML]{3531FF} {67.38}}} & {\color[HTML]{3531FF} {62.98}}          \\ \hline
Ours(EW)          & {\color[HTML]{333333} 70.33}         & \multicolumn{1}{c|}{{\color[HTML]{333333} 61.00}}          & \multicolumn{1}{c|}{{\color[HTML]{3531FF} 46.67}}          & \multicolumn{1}{c|}{{\color[HTML]{333333} 44.10}} & {\color[HTML]{333333} 60.83}          & \multicolumn{1}{c|}{{\color[HTML]{333333} \textbf{67.56}}} & {\color[HTML]{333333} \textbf{63.15}}          \\ \hline
Ours(DW)          & {\color[HTML]{333333} 70.30}         & \multicolumn{1}{c|}{{\color[HTML]{333333} 60.96}}          & \multicolumn{1}{c|}{{\color[HTML]{333333} 46.63}}          & \multicolumn{1}{c|}{{\color[HTML]{333333} 44.03}} & {\color[HTML]{333333} 60.79}          & \multicolumn{1}{c|}{{\color[HTML]{333333} \textbf{67.56}}} & {\color[HTML]{333333} \textbf{63.15}}          \\ \hline
\end{tabular}
\end{table}

KNN-Defense evidently outperforms other 3D defense methods against untargeted point-dropping attacks \cite{zheng2019pointcloud} on ModelNet40. It also achieves state-of-the-art results against most targeted attacks on PointNet \cite{qi2017pointnet} and PCT \cite{qi2017pointnet++}, while maintaining reasonable performance in classifying clean images across different architectures.

Table \ref{table-pct-scan} presents numerical results demonstrating the effectiveness of the proposed defense method against adversarial attacks on PCT model using the ScanObjectNN dataset. It can be observed from this table that although KNN-Defense does not consistently outperform all other defenses, it still delivers competitive results, particularly against untargeted attacks. One of the challenges posed by this dataset is the model’s limited ability to generate highly discriminative features, as reflected by the baseline accuracy being below 77\%. This limitation appears to have some impact on the performance of KNN-Defense. Nevertheless, the method continues to perform reasonably well, suggesting a level of robustness even in the presence of suboptimal feature representations.

\subsubsection{Inference Efficiency}

Table~\ref{time_table} reports the average inference time per point cloud for different defense methods across all four classification architectures. Among the evaluated techniques, SOR and IF-Defense exhibit the highest computational cost, with average runtimes of 40.32 ms and 31.05 ms, respectively. These methods rely on geometric operations or internal model computations, which can hinder deployment in real-time systems.
In contrast, the proposed KNN-Defense with uniform weighting (UW) achieves a significantly lower runtime of 5.44 ms—approximately 6× faster than IF-Defense and over 7× faster than SOR—while maintaining strong adversarial robustness. Although SRS is the fastest method at 1.57 ms, it shows inferior defense performance in earlier evaluations (Tables~1–5).
These findings highlight the trade-off offered by KNN-Defense between inference speed and robustness, making it suitable for time-sensitive or resource-constrained applications

\begin{table}[t!]
\captionsetup{width=\textwidth, justification=justified}
\caption{Average inference runtime per point cloud for different defense methods, computed across four classification models.}\label{time_table}
\begin{tabular}{|c|c|}
\hline
\textbf{Defenses}       & \textbf{Time (milliseconds)} \\ \hline
SRS            & 1.57               \\ \hline
SOR            & 40.32              \\ \hline
DUP-Net        & 7.99              \\ \hline
If-Defense     & 31.05              \\ \hline
Ours (Uniform) & 5.44              \\ \hline
\end{tabular}
\end{table}

\section{Conclusion}

This paper proposed a defense framework for 3D point cloud classification that improved the robustness of 3D models by leveraging the manifold assumption and semantic neighborhood relations in the feature space. Rather than reconstructing surface geometry or applying geometric priors, the method restored the perturbed adversarial examples based on feature-space similarity with training instances. A key advantage of this method was its alignment with the structure of learned feature representations. By operating on intermediate features extracted from pretrained models, the method eliminated the need for architectural changes or retraining. Its lightweight design ensured fast inference, making it highly suitable for real-time and embedded 3D perception systems. Furthermore, its generality across attack types allowed for effective defense without tailoring to specific threat models. Experimental evaluations confirmed that the proposed method outperforms the existing 3D defense methods, particularly under point-dropping and point-shifting attacks—where conventional approaches often fail due to the selective removal of critical structural points. Those findings underscored the importance of manifold-based semantic alignment in developing scalable and resilient defenses for 3D vision systems.

\bibliography{sn-bibliography}

\end{document}